\documentclass[amsfonts, amssymb, amsmath, reprint]{revtex4-2}

\usepackage{graphicx, subcaption}
\usepackage{tikz, standalone}
\usepackage{pgfplots}
\usepgfplotslibrary{colormaps} 
\pgfplotsset{compat=newest}
\usepackage{upgreek}
\usepackage{lipsum}
\usepackage{comment}
\usepackage{nicefrac}
\newcommand{\um}{\upmu\textrm{m}}
\newcommand{\spatial}{\textsf{S}}
\newcommand{\temporal}{\textsf{T}}
\newcommand{\combined}{\textsf{C}}

\usetikzlibrary{fit, backgrounds}
\usetikzlibrary {fadings}

\DeclareMathOperator{\arctantwo}{arctan2}

\begin{document}

\title{Learning optimal integration of spatial and temporal information in noisy chemotaxis}
\author{Albert Alonso}
\author{Julius B. Kirkegaard}
\email[Correspondence email address: ]{julius.kirkegaard@nbi.ku.dk}
\affiliation{Niels Bohr Institute, University of Copenhagen, Denmark}
\affiliation{Department of Computer Science, University of Copenhagen, Denmark}

\date{\today}

\begin{abstract}
    \textbf{Abstract:} We investigate the boundary between chemotaxis driven by spatial estimation of gradients and chemotaxis driven by temporal estimation.
    While it is well known that spatial chemotaxis becomes disadvantageous for small organisms at high noise levels, it is unclear whether there is a discontinuous switch of optimal strategies or a continuous transition exists.
    Here, we employ deep reinforcement learning to study the possible integration of spatial and temporal information in an a priori unconstrained manner.
    We parameterize such a combined chemotactic policy by a recurrent neural network and evaluate it using a minimal theoretical model of a chemotactic cell.
    By comparing with constrained variants of the policy, we show that it converges to purely temporal and spatial strategies at small and large cell sizes, respectively.
    We find that the transition between the regimes is continuous, with the combined strategy outperforming in the transition region both the constrained variants as well as models that explicitly integrate spatial and temporal information.
    Finally, by utilizing the attribution method of integrated gradients, we show that the policy relies on a non-trivial combination of spatially and temporally derived gradient information in a ratio that varies dynamically during the chemotactic trajectories.
\end{abstract}

\maketitle

\section{Introduction}

Chemotaxis, the directed motion of organisms towards or away from chemical cues, is a fundamental biological mechanism that spans biological kingdoms.
For instance, prokaryotes rely on chemotaxis to find nutrients, avoid toxins or even optimize oxygen and pH levels by sensing molecular cues \cite{biStimulusSensingSignal2018, matillaCatalogueSignalMolecules2022, matillaEffectBacterialChemotaxis2018}.
Single-celled eukaryotes show similar chemotactic traits \cite{willardSignalingPathwaysMediating2006}, and countless biological processes in multicellular eukaryotes are supported by chemotaxis such as the fighting of bacterial infections by white blood cells, the positioning of stem cells during early embryonic development, and formation of multicellular structures in slime mould development \cite{yussofCellularEventsBiomarkers2012,christensenCirculationChemotaxisFetal2004,willardSignalingPathwaysMediating2006}. 
Likewise, a hallmark of cancer metastasis is the chemotaxis of tumour cells towards blood vessels \cite{hunterMechanismsMetastasis2008}.

\begin{figure*}[thpb]
    \begin{subfigure}[c]{1.7in}
        \caption{}
        \definecolor{cell_color}{HTML}{619B8A}
\definecolor{sensor_color}{HTML}{FE7F2D}
\begin{tikzpicture}

\tikzstyle{sensor} = [draw=cell_color!50!sensor_color]
\tikzstyle{cell} = [draw=black, very thick, fill=cell_color!10]
\tikzstyle{receptor} = [draw=black, very thick, fill=cell_color]
\tikzstyle{molecule} = [fill=orange!50, opacity=1]

\foreach \i in {180,...,0}{
    \draw[molecule] (0, 0)+(\i:rand*3.5) circle (0.05cm);
}

\draw[cell] (0,0) circle (2cm);
\draw[very thick, dashed] (0,0) -- +(72:2);
\draw[very thick, dashed] (0,0) -- +(0:2);
\foreach \i [evaluate=\i as \x using \i/2] in {0,72,...,288} {
    \draw[receptor, fill=cell_color!\x] (\i:2cm) circle (0.15cm);
    \draw[sensor] (\i:2cm) circle (1.17cm);
}
\draw[very thick] (0.5,0) arc (0:72:0.5) node[midway, above, right, scale=1.8] {$\alpha$};

\draw[very thick, latex-latex] (0,0) -- +(180:2.0) node[above, midway, sloped, very thick, scale=2, rounded corners, inner sep=0.5] {$R$};
\draw[very thick, latex-latex] (72+72:2) -- +(-10:-1.17) node[above, midway, sloped,  inner sep=1, thick, scale=1.5] {$r_s$};

\draw[-latex, ultra thick, red!30!gray] (-2, 3.5) -- (2, 3.5) node [midway, above] {Swimming direction};

\end{tikzpicture}
        \label{fig:model-diagram}
    \end{subfigure}
    \begin{subfigure}[c]{2.6in}
         \caption{}
        \definecolor{cell_color}{HTML}{619B8A}
\definecolor{sensor_color}{HTML}{FE7F2D}
\definecolor{path}{HTML}{326dba}
\definecolor{border}{HTML}{2A9D8F}

\begin{tikzpicture}
\pgfplotsset{/pgfplots/colormap={Oranges}{
rgb255=(255, 245, 235)
rgb255=(254, 230, 206)
rgb255=(253, 208, 162)
rgb255=(253, 174, 107)
rgb255=(253, 141, 60)
rgb255=(241, 105, 19)
rgb255=(217, 72, 1)
rgb255=(166, 54, 2)
rgb255=(127, 39, 4)
    }}
\tikzstyle{annotation} = [text=black]

\tikzstyle{cell} = [fill=cell_color!50, draw=black]
\tikzstyle{receptor} = [fill=sensor_color!50, draw=black]
\tikzstyle{sensor} = [draw=none, fill=sensor_color, fill opacity=0.3]
\tikzstyle{molecule} = [fill=red!30!yellow!70, opacity=0.8]

\begin{axis}[
    draw=none,
    view={0}{90},
    grid=none,
    hide axis,
]
    \addplot3[
        patch,
        domain=10:-80,
        samples=10,
        shader=interp,
        grid=none,
    ]  {exp(-sqrt(0.01/10)*sqrt(x^2 + y^2))}; 

\end{axis}

\begin{scope}[shift={(3, 2)}, scale=0.1]
\draw[path, ->, rounded corners=6, very thick, path fading=west] (-5, -4) -- (-3,0) -- (2, 0) -- ++(3.5, 0) -- ++(0, 10) -- ++(10, 3) -- ++(3, 4);
\draw[->, rounded corners=6, annotation, thick] (5,-2) -- ++(2, 0) node [midway, below, anchor=west, scale=0.9, annotation] {$\Delta \theta$} -- ++(0, 2.5) ;

\begin{scope}[shift=({-6, -6}), rotate=70, opacity=0.2]
\pgftransparencygroup
\draw[cell] (0,0) circle (2);
\foreach \i [evaluate=\i as \x using \i/2] in {0,72,...,288} 
    \draw[receptor, fill=cell_color!\x] (\i:2) circle (0.5);
\endpgftransparencygroup
\end{scope}

\begin{scope}[opacity=0.4]
\pgftransparencygroup
\draw[cell] (0,0) circle (2);
\foreach \i [evaluate=\i as \x using \i/2] in {0,72,...,288} 
    \draw[receptor, fill=cell_color!\x] (\i:2) circle (0.5);
\endpgftransparencygroup
\end{scope}

\begin{scope}[shift=({5.5, 5}), rotate=90, opacity=0.6]
\pgftransparencygroup
\draw[cell] (0,0) circle (2);
\foreach \i [evaluate=\i as \x using \i/2] in {0,72,...,288} 
    \draw[receptor, fill=cell_color!\x] (\i:2) circle (0.5);
\endpgftransparencygroup
\end{scope}

\begin{scope}[shift=({9, 11}), rotate=20, opacity=0.8]
\pgftransparencygroup
\draw[cell] (0,0) circle (2);
\foreach \i [evaluate=\i as \x using \i/2] in {0,72,...,288} 
    \draw[receptor, fill=cell_color!\x] (\i:2) circle (0.5);
\endpgftransparencygroup
\end{scope}

\begin{scope}[shift=({15.7, 13.7}), rotate=45, opacity=1]

\draw[cell] (0,0) circle (2);
\foreach \i [evaluate=\i as \x using \i/2] in {0,72,...,288} 
    \draw[receptor, fill=cell_color!\x] (\i:2) circle (0.5);
\end{scope}

\end{scope}

\node[annotation, anchor=north west, align=center, scale=1.2] (source_label) at (1, 5) {Chemoatractant\\source};
\node[circle, anchor=north west, green] (source) at (5.4, 5.2) {};
\draw[-latex, annotation, very thick] (source_label) -- (source); 

\end{tikzpicture}
         \label{fig:simulation}
    \end{subfigure}
    \begin{subfigure}[c]{2.0in}
        \caption{}
        \includegraphics[width=\linewidth]{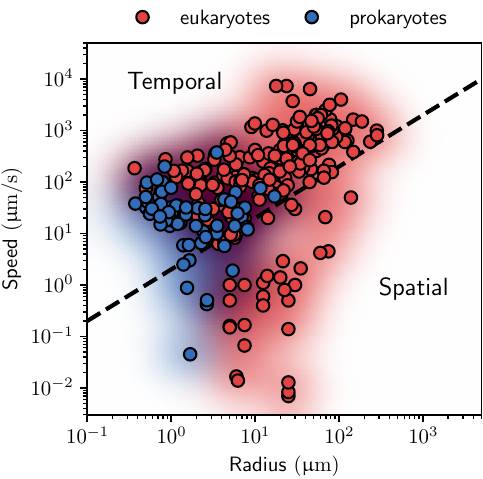}
        \label{fig:phase-speed-size}
    \end{subfigure}
    \begin{subfigure}[c]{4.5in}
        \caption{}
        \definecolor{color1}{HTML}{264653}
\definecolor{color2}{HTML}{e24444}
\definecolor{color3}{HTML}{326dba}
\definecolor{color4}{HTML}{F4A261}

\begin{tikzpicture}[
thick, scale=0.6, rounded corners=5,
]
\begin{scope}[shift={(8.2, 0)}]
    \tikzstyle{value}[color1!20] = [fill=#1, inner sep=1, outer sep=1]
    \tikzstyle{network} = [draw=color1!50!black, fill=color1!5, align=center, inner sep=6,  rounded corners=0]
    \tikzstyle{connector}[color1]=  [color=#1, -latex]
    
    \node[value=none] (m) at (4, 10.5) {$\{m_t^{(1)}, m_t^{(2)}, ..., m_t^{(K)}\}$};
    \node[value=none] (h_prev) at (0, 8) {};
    \node[value=none] (h_next) at (8.0, 8) {};
    \node[value=none] (a_prev) at (-0.0, 7) {};
    
    \node[network] (recurrent) at (4, 8) {GRU};
    \draw[connector] (m) -- (recurrent);
    
    \node[network] (policy) at (5.5, 5) {MLP};
    \node[network] (critic) at (2.5, 5) {MLP};
    
    \node[value, fill=color1!20] (ht) at (4, 6.5) {$h_t$};
    
    \draw[connector](recurrent) -- (ht) -| (policy);
    \draw[connector](recurrent) -- (ht) -| (critic);
    
    \draw[connector] (h_prev) -- (recurrent) node[pos=0.5, above, style=value, text=black] {$h_{t-1}$};
    \draw[connector] (recurrent) -- (h_next) node[pos=0.5, above, style=value, text=black] {$h_{t}$};
    \draw[connector, color4] (a_prev) -- ++(2.5, 0) node[pos=0.6, above, style=value, text=black] {$a_{t-1}$}  -- ++(0, 0.6) -- (recurrent) ;
    
    \node[value=white] (action) at (5.5, 2.5) {$a_t \sim \mathcal{N}(\mu_t, \sigma_t)$};
    \draw[connector] (policy) -- (action);
    
    \node[value=white] (value) at (2.5,2.5) {$V_t$};
    \draw[connector] (critic) -- (value);
    
    \draw[color4] (action) -- ++(2.2,0) -- ++(0, 4.5) -- ++(0.3, 0);
    
    \begin{scope}[on background layer]
        \node[thick, draw=none, fit=(recurrent)(policy)(critic), inner sep=13, fill=color1!20, label={[anchor=north west, text=black, font=\bfseries]north west:Combined}] {};
    \end{scope}
\end{scope}

\begin{scope}[shift={(-7.8, 0)}]
    \tikzstyle{value} = [fill=white, inner sep=1, outer sep=1]
    \tikzstyle{network} = [draw=color2!50!black, fill=color2!5, align=center, inner sep=6, rounded corners=0]
    \tikzstyle{connector}[color1]=  [color=#1, -latex]
    \node[value] (m) at (4, 10.5) {$\{m_t^{(1)}, m_t^{(2)}, ..., m_t^{(K)}\}$};
    
    \node[network] (recurrent) at (4, 8) {Dense};
    \draw[connector] (m) -- (recurrent);
    
    \node[value, fill=color2!20] (ht) at (4, 6.5) {$h_t$};
    
    \node[network] (policy) at (5.5, 5) {MLP};
    \node[network] (critic) at (2.5, 5) {MLP};
    
    \draw[connector](recurrent) -- (ht) -| (policy);
    \draw[connector](recurrent) -- (ht) -| (critic);
    
    \node[value] (action) at (5.5, 2.5) {$a_t \sim \mathcal{N}(\mu_t, \sigma_t)$};
    \draw[connector] (policy) -- (action);
    
    \node[value] (value) at (2.5, 2.5) {$V_t$};
    \draw[connector] (critic) -- (value);
    
    \begin{scope}[on background layer]
        \node[thick, draw=none, fit=(recurrent)(policy)(critic), inner sep=13, fill=color2!20, label={[anchor=north west, text=black, font=\bfseries]north west:Spatial}] {};
    \end{scope}
\end{scope}

\begin{scope}[shift={(0.0, 0)}]
    \tikzstyle{value}[color3!20] = [fill=#1, inner sep=1, outer sep=1]
    \tikzstyle{network} = [draw=color3!50!black, fill=color3!5, align=center, inner sep=6,  rounded corners=0]
    \tikzstyle{connector}[color1]=  [color=#1, -latex]
    
    \node[value=none] (m) at (4, 10.5) {$\langle \vec{m}_t \rangle$};
    \node[value=none] (h_prev) at (0, 8) {};
    \node[value=none] (h_next) at (8, 8) {};
    \node[value=none] (a_prev) at (0, 7) {};
    
    \node[network] (recurrent) at (4, 8) {GRU};
    \draw[connector] (m) -- (recurrent);
    
    \node[network] (policy) at (5.5, 5) {MLP};
    \node[network] (critic) at (2.5, 5) {MLP};
    
    \node[value] (ht) at (4, 6.5) {$h_t$};
    
    \draw[connector](recurrent) -- (ht) -| (policy);
    \draw[connector](recurrent) -- (ht) -| (critic);
    
    \draw[connector] (h_prev) -- (recurrent) node[pos=0.5, above, style=value, text=black] {$h_{t-1}$};
    \draw[connector] (recurrent) -- (h_next) node[pos=0.5, above, style=value, text=black] {$h_{t}$};
    \draw[connector, color2] (a_prev) -- ++(2.5, 0) node[pos=0.6, above, style=value, text=black] {$a_{t-1}$}  -- ++(0, 0.6) -- (recurrent) ;
    
    \node[value=white] (action) at (5.5, 2.5) {$a_t \sim \mathcal{N}(\mu_t, \sigma_t)$};
    \draw[connector] (policy) -- (action);
    
    \node[value=white] (value) at (2.5,2.5) {$V_t$};
    \draw[connector] (critic) -- (value);
    
    \draw[color2] (action) -- ++(2.2,0) -- ++(0, 5) -- ++(0.3, 0);
    
    \begin{scope}[on background layer]
        \node[thick, draw=none, fit=(recurrent)(policy)(critic), inner sep=13, fill=color3!20, label={[anchor=north west, text=black, font=\bfseries]north west:Temporal}] {};
    \end{scope}
\end{scope}
\end{tikzpicture}  
         \label{fig:policy}
    \end{subfigure}
    \caption{
    \textbf{A}~Representation of the model cell with five sensors surrounded by chemoattractant particles. Each sensor measures the number of particles $M_i$ inside its sensing range $r_s$ and transforms it as $m_i=log(M_i+1)$.
    \textbf{B}~Illustration of the simulation environment where the cell navigates towards the centre of the chemoattractant source.
    \textbf{C}~Phase space diagram of cell sizes and speeds showing the distribution of common unicellular prokaryotes and eukaryotes.
    The dashed line roughly indicates the binary division between temporal and spatial navigation strategies \cite{wanOriginsEukaryoticExcitability2021}.
    Data from Refs. \cite{wanOriginsEukaryoticExcitability2021, rodriguesBankSwimmingOrganisms2021}.
    \textbf{D}~Our three neural network policies output the cell's action based on the measurements and hidden states. The combined policy has access to the individual measurement of its sensors and has a hidden state used in a recurrent neural network layer, whereas spatial and temporal only have one of these features.
    \textit{Dense}: a linear NN layer connecting all inputs with all outputs. \textit{MLP}: Multilayer perceptron, a sequence of dense layers with non-linear activations. \textit{GRU}: Gated recurrent unit, a simple form of recurrent neural network module, which combines a hidden state with new input.
    The policy output of the model is both a mean value $\mu_t$ and a standard deviation $\sigma_t$, which defines a normal distribution from which an action $a_t$ is sampled. In our experiments, $\sigma_t \rightarrow 0$ at the end of training (see SI) results in deterministic policies.
    }
\end{figure*}

However, the ubiquity of chemotaxis in biology does not imply uniformity in the mechanisms that underlie the navigation.
At the scale of microorganisms, the fluctuations of the molecules that bind to the cells' receptors are non-negligible and impose physical limits on the accuracy of the measurements and, thus navigation.
Chemotaxis is typically dichotomized into \textit{spatial} and \textit{temporal} strategies (FIG~\ref{fig:phase-speed-size}) \cite{wanOriginsEukaryoticExcitability2021, tanComputationalModelHow2018, dusenberySpatialSensingStimulus1998}.
Larger cells, usually eukaryotes, primarily exploit spatial sensing, harnessing their size to directly perceive chemical concentration gradients \cite{endresAccuracyDirectGradient2008},
whereas smaller cells like bacteria are known to adopt temporal sensing, detecting alterations in chemical concentrations temporally to deduce information on the gradient's direction, as the fluctuations across their body render spatial sensing useless \cite{bergMotileBehaviorBacteria2000, sourjikRespondingChemicalGradients2012, bialekPhysicalLimitsBiochemical2005}.
These differences in sensing mechanisms have direct consequences for the possible types of navigation decision processes.

This binary classification enables detailed analysis of the distinct forms of chemotaxis within each category.
However, as the optimal strategy is dependent on continuously varying parameters such as the size and velocity of the organism as well as the chemoattractant concentration, it leaves the question of whether organisms can utilize an integration of both spatial and temporal sensing mechanisms in their chemotactic strategies \cite{delisiTheoryMeasurementError1983}, and whether such a combination would be preferential in intermediate ranges of these parameters.
Interestingly, it has been shown that cells thought only to use spatial sensing also rely on temporal information during chemotaxis when given periodic waves of chemoattractant  \cite{karmakarCellularMemoryEukaryotic2021,nakajimaRectifiedDirectionalSensing2014}.
Static temporal averaging of previous measurements has also been shown to reduce sensing noise on cells placed in shallow concentrations \cite{endresAccuracyDirectGradient2008}; however, this does not take into account the effect of the motile cell itself reacting to the measurements.
Previous work has proposed a more complex inclusion of both types of sensing to develop newer strategies without being able to outperform single sensing strategies \cite{metznerEfficiencyChemotacticPursuit2019}, showcasing that efficient integration of both strategies is probably non-trivial.

Here, we employ deep reinforcement learning (DRL) to discover optimal chemotactic strategies that can combine spatial and temporal sensing.
Previous work has successfully made use of DRL to find optimal strategies for self-propelled agents exploiting the flow in fluid environments \cite{colabreseFlowNavigationSmart2017} and for studying the tracking policies of flying insects relying on memory from noisy measurements to locate food or other insects \cite{singhEmergentBehaviourNeural2023}.
Similarly, machine learning has been utilized for demonstrating the optimality of known chemotactic strategies \cite{hartlMicroswimmersLearningChemotaxis2021, ramakrishnanLearningRunandtumbleChemotaxis2023}.

We propose a minimal single-cell model and use modern policy optimization techniques \cite{schulmanProximalPolicyOptimization2017} to identify the strategy that minimises the time it takes the cell to reach a source of chemoattractant.
The model cell is endowed with distinct sensors that enable spatial gradient estimation and is given an internal memory state that allows temporal information to be derived. Based on a combination of these inputs, the cell must modify its orientation, a mapping that we leave largely unconstrained by employing deep neural networks.

We demonstrate the existence of a better performant chemotactic strategy that non-trivially combines spatial and temporal sensing.
Specifically, we pinpoint a range of cell sizes where a combined sensing strategy outperforms optimal single sensing strategies.
We then concentrate our analysis on this interface, comparing it to analytical ones and offering both qualitative and quantitative insights on the internal dynamics of the optimal navigational policy.

\section{\label{s:methods}Methods}

\subsection{\label{ss:simulation}The simulation model}

We study a exponentially decaying, two-dimensional distribution $C(\boldsymbol{x})$ of chemoattractant particles with a concentration peak at $\boldsymbol{x} = 0$,
\begin{equation}
    C(\boldsymbol{x}) = C_0 \exp \left( -\lambda \, |\boldsymbol{x}| \right).
\end{equation}
We take $\lambda = 0.01 \upmu\mathrm{m}^{-1}$, and study $C_0$ varying from $C_q = 16 \upmu\mathrm{m}^{-2}$ to $10 \cdot C_q$, which in turn sets the signal-to-noise ratio of the system.
In the SI, we further give examples of algebraic and Bessel function concentration profiles
which can e.g. arise from decaying and diffusing particles emanating from a central static source,
\begin{equation} \label{eq:concentration-eq}
    D \, \nabla^2 C(\boldsymbol{x}) - \kappa \, C(\boldsymbol{x}) + \rho \, \delta(\boldsymbol{x}) = 0.
\end{equation}
Here, $D$ is the particle diffusion coefficient, $\kappa$ is a particle decay rate, which sets a length scale $\sqrt{D/\kappa}$.
$\rho$ is the rate of particle release at $\boldsymbol{x} = 0$.

Our cell model consists of a circular disk of radius $R$, equipped with $K$ sensors uniformly spaced around its surface (FIG.~\ref{fig:model-diagram}), whose objective is to reach the source of the chemoattractant by controlling its direction of motion depending on the environmental measurements.
The cell senses the environment through molecules binding to cell-surface receptors. Still, in the interest of keeping our model as simple as possible, we neglect the complex receptor dynamics of receptor binding and unbinding \cite{bergPhysicsChemoreception1977}.
Thus, we assume each of the $K$ sensors to possess a detection area of radius $r_s = R \, \sin(\pi / K)$ such that the entire surface of the cell is covered.
We fix $K = 5$ for all our experiments.

Our model cell moves forward at a constant speed $v$, with its trajectory orientation $\theta(t)$ being modified both by its own actions as well as due to rotational noise,
\begin{equation}
    \mathrm{d}{\theta} = a_t \, \mathrm{d}t + \sqrt{2 D_R} \, \mathrm{d} W.
\end{equation}
Here, $a_t$ is the output of the cell's navigational policy $\pi$, and the second term is Wiener noise with rotational diffusion coefficient $D_R$.
Rotational diffusion forces the cells' navigation policies to react to the sensor signals at least on a time scale $1/D_R$ \cite{celaniBacterialStrategiesChemotaxis2010}.
In our experiments we use $v = 5 \, \upmu \mathrm{m}/\mathrm{s}$ and $D_r = 0.025 / \mathrm{s}$, and use a time-stepping of $\Delta t = 0.1 \, \mathrm{s}$ to solve the stochastic equations.

We model the cell receptors as \textit{perfect instruments} \cite{bergPhysicsChemoreception1977}, and thus, at each time step of our simulation, the sensors measure the number of molecules inside their sensor range.
This induces fluctuations in measurements with a signal-to-noise ratio that increases with concentration.
Thus, nutrient-deprived environments with a low number of detected particles are noisy, and nutrient-rich environments are more deterministic.

We approximate the particle count within each sensor's area as a stochastic process sampling from a Poisson distribution.
Exploiting the nearly constant particle density over the detection area, we use
\begin{align}
    \mathbb{E}(M_i) &= \int_A C(\boldsymbol{x}) \, \mathrm{d} A \approx C(d_i) \cdot \pi r_s^2  \label{eq:average-particles} \\ 
    M_i &\sim \mathrm{Poisson}(\mathbb{E}(M_i)) \label{eq:sampled-particles},
\end{align}
$d_i$ being the radial distance of the receptor centre to the source of the chemoattractant.

Simulations are initialized at random distances $d_0$ from the source with random orientations $\theta_0$ and crucially with a rate of particle release $\rho$, which we sample in the range $\rho_0$ and $10 \cdot \rho_0$.
These random initializations ensure that the cell agents cannot overtrain to specific molecule counts and specific trajectories but rather need to generalize across noise levels and become adaptable to varying concentration profiles.
Specifically,

Finally, we do biologically inspired preprocessing of the receptor input by transforming according to the Weber–Fechner law \cite{kalininLogarithmicSensingEscherichia2009},
\begin{equation}
    \label{eq:weber-fechner}
    m_i = \log \! \left(M_i + 1\right).
\end{equation}
While this could have been learned directly from the data, it conveniently brings the neural network input to a tightly constrained domain that is more suitable for DRL, and also means that noise in $m_i$ decreases not just relative to the signal but also in absolute numbers as $\rho$ increases.

\subsection{\label{sec:policy}The policy}

The internal mechanisms of a chemotactic cell involve a complex set of biochemical spatio-temporal reactions.
Here, we do not model these reactions explicitly, but instead model directly an input-output approximator, \textit{the cell policy} $\pi$.
This policy maps an internal state $s_t$, which in the simplest case could just be the vector of instantaneous measurements, to an action $a_t$.
We parameterize the function by using artificial neural networks (ANN) to minimize expressive restrictions on the learned policy.

To estimate the cell policy, we assume that it is an optimizer of efficient chemotaxis, which we define as minimizing the time it takes to reach a certain distance from the source.
More precisely, at the end of a simulation, we calculate a \textit{reward} by
\begin{equation}
    \mathcal{R} = \frac{t_{max} - \tau}{t_{max}} + \max(-1, \frac{\delta - d}{d_0 - \delta}),
\end{equation}
where $d$ is the final distance to the source (which will be $=\delta$ if the source has reached) and $\tau$ is the time it took to reach the source (or $=t_{max}$ if the source was not reached).
The first term is a normalized reward for getting to the source fast and the second is a bootstrapping reward that punishes cells that do not reach within the required distance $\delta$ of the source.
The reward is normalized between $[-1, 1]$, as is convention in reinforcement learning.
Simulations terminate when the cell has reached $\delta$ distance to the source or the simulation time has exceeded $t_{\mathrm{max}}$, thus only one term of the reward expression is nonzero at the end of the episode;
with the distance reward dominating early in training and the time reward at the end of training.

To find the optimal ANN policy, we employ Proximal Policy Optimization (PPO) \cite{schulmanProximalPolicyOptimization2017}, which adapts the policy $\pi$ in order to maximize the average reward.
We study three variants of the agents (FIG.~\ref{fig:policy}):
one policy we restrict to act purely on instantaneous \textit{spatial} information.
This is enforced by simply designing the neural network to be a pure feedforward network --- from measurements $\{m^{(1)}_t, m^{(2)}_t, \cdots, m^{(K)}_t\}$ to output $a_t$.
Likewise, we design a purely \textit{temporal} network, which does not receive spatial information but rather the average of all receptors $\langle m_t \rangle$.
Instead, this agent must rely on memory to provide temporal information on the particle gradients.
This is achieved by introducing a recurrent layer into the policy neural network, which emulates the biochemical memory of real cells.
Finally, we study a \textit{combined} agent, which has access to both spatially resolved measurements and has memory that can be used to derive temporal information.
This agent can execute pure spatial and pure temporal strategies but can furthermore act on any combination of this information.
Network details are given in SI.

Our networks also output an estimate of the final reward $V_t$ (FIG.~\ref{fig:policy}), which the PPO algorithms use to speed up convergence, but which does not influence the policy once trained.
Further, as the nature of PPO's exploration strategy adds noise to the policy output, we also recurrently feed the cell's action back into the temporal policies, which aids the training in reaching a deterministic strategy without hindering stochastic exploration.

\section{\label{s:results}Results}

\subsection{\label{ss:noise}Optimizing for noise-robust strategies}

\begin{figure}[ht]
    \centering
    \includegraphics[width=\linewidth]{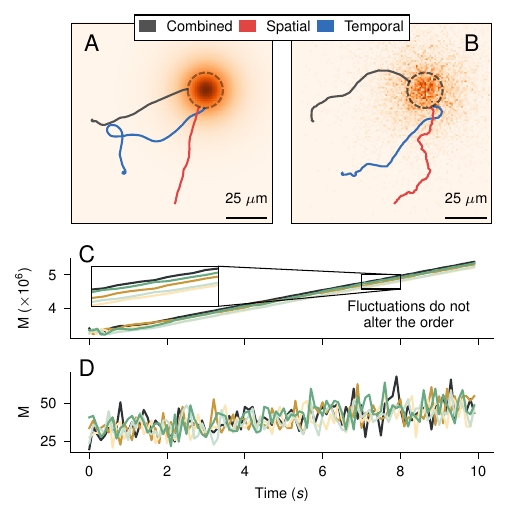}
    \caption{
    \textbf{A-B}~Example trajectories of found strategies at $R=2\um$ for each variant, in nutrient-rich media ($C_0/C_q = 10^4$) and at nutrient-depleted environments ($C_0/C_q=1$), respectively.
    Circles indicate $\delta = 10 \, \upmu \mathrm{m}$.
    \textbf{C-D}~Measurement values of each sensor of the cell at \textbf{A} and \textbf{B} respectively. Each color represents one of the $K=5$ sensors used in the trajectories. The measurements correspond to those of the Combined cell.
    }
    \label{fig:visual-trajectories}
\end{figure}

Our deep reinforcement learning approach is designed principally to work at all noise levels. 
In nutrient-rich environments, where input to the agents are not corrupted by nosie, our DRL framework converges quickly to effective temporal and spatial strategies.
Resulting trajectories in these environments are close to deterministic as the noise from measurements gets reduced, and fewer mistakes in orientation corrections tend to occur.
In those scenarios, spatial-based gradient estimation is effective in directly locating the source of chemoattractants and noise due to rotational diffusion does not pose a challenge for the cell, which only needs to follow the strength of the sensors (FIG.~\ref{fig:visual-trajectories}C).
Likewise, the optimal temporal sensing strategy at high concentrations is easily understood as it continuously measures the change in concentration and increases the turn when the concentration starts diminishing.
As the temporal strategy contains no information about the sensors' positions, it has to spontaneously break its rotational symmetry, which is exemplified in the resulting left-turning shown in FIG.~\ref{fig:visual-trajectories}A, resembling e.g. the chirality of sperm chemotaxis trajectories \cite{alvarezComputationalSpermCell2014}.

In contrast, in the low concentration limit, the input to the cell receptors is extremely noisy (FIG.~\ref{fig:visual-trajectories}D), and the identification of optimal strategies becomes less clear.
Yet, our DRL approach is able to identify working strategies both using purely spatial and purely temporal sensing mechanisms (FIG.~\ref{fig:visual-trajectories}B).
Qualitatively, we note that the identified low-concentration temporal strategy behaves very robustly against noise, as its trajectory remains smooth despite its stochastic input.
This can be interpreted as \textit{low reactivity}, which also showcases itself as the temporal strategy only slowly adapts its trajectory as it nears the source.
In comparison, the spatial strategy is very reactive, and while this makes it susceptible to the stochastic input, it enables it to quickly adapt its orientation once it nears the source and the concentration is relatively high.
Finally, we observe the first hint that the combined strategy can outperform the two: it shows both low reactivity when it is far from the source and high reactivity once in its proximity.

While deterministic policies are fast to identify, the information which reinforces policies in the low concentration limit is much more stochastic, making the optimization process harder.
To enable learning in this very noisy regime, our reinforcement learning steps rely on averaging the result of thousands of runs and require millions of simulations to converge to a solution (see SI table).
To make this feasible, we developed a custom end-to-end RL implementation which runs exclusively on GPUs (see Code Availability).

We note that DRL is in not guaranteed to find the globally optimal policy.
However, we find that independent runs of the DRL training procedure result in the same policies, which hints that the obtained local optima could be global.

\subsection{Smooth transition between a temporal and a spatial strategy}

For evaluation, we define a strategy's chemotactic efficiency $\eta$ by how fast the cell reaches the source compared to the minimal time a cell of speed $v$ would take to reach it from the same initial position (note that this is independent of $t_{max}$ which was used for training).
Thus, the efficiency of a strategy is given by
\begin{equation}
    \eta = \left\langle \dfrac{d_0-\delta}{v \cdot \tau} \right\rangle,
\end{equation}
where $\tau$ is the time it takes the cell to reach the source threshold distance $\delta$ and $d_0$ is the initial distance to the source, and the average is taken over all realisations.

\begin{figure}[!ht]
    \centering
    \includegraphics[width=\linewidth]{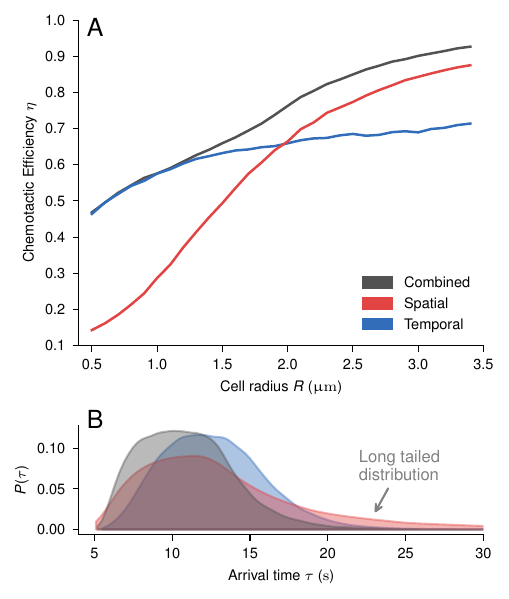}
    \caption{
    \textbf{A}~Chemotactic efficiency of each variant on reaching the source as a function of cell size.
    Each value is the result of training and evaluating the policies at that cell radius for sampled values of $C_0$.
    The average efficiency is evaluated on $2^{16}$ independent runs.
    A ``blind'' agent obtains efficiency $\eta \approx 0.02$.
    \textbf{B}~Distribution of arrival times to the source of the three cell variants at $R=2\um$. All evaluations use sampled concentrations.
    }
    \label{fig:performance-radius}
\end{figure}

We train our three variants, spatial (\spatial{}), temporal (\temporal{}) and combined (\combined{}), on the same simulation parameters at different cell sizes and proceed to calculate their efficiencies (FIG.~\ref{fig:performance-radius}A).
At small sizes, where the positional sensor information becomes indistinguishable due to the noise, both \temporal{} and \combined{} policies show the same performance.
This is in accordance with previous studies showing that small cells are incapable of sensing gradients along their own body due to the fluctuations in measurements \cite{delisiTheoryMeasurementError1983}.
Nevertheless, as the cell size increases, \combined{} starts to outperform \temporal{}, indicating that the tiny amount of available gradient information, as observed by the poor performance of \spatial{}, can somehow be integrated into a temporally dominated strategy to improve its performance.
At large cell sizes, \spatial{} dominates \temporal{}, and while a gap still remains between \spatial{} and \combined{} at large $R$, it shows convergence towards the same strategy.
Thus, at the largest scales, the sensors need not rely heavily on old measurements to accurately estimate the gradient.
At intermediate cell sizes, we find that the optimal strategy is not purely spatial or temporal.
In detail, we observe a smooth transition between strategies, indicating that there is a continuous integration of information stemming from spatial input and memory.
Despite being dominated by noise, as illustrated in FIG.~\ref{fig:visual-trajectories}D, \combined{} is capable of taking advantage of the measurement differences between the different receptors on the cell surface to improve its efficiency.
To explore this integration, we now focus on this intermediate region where both \spatial{} and \temporal{} perform similarly yet are outperformed by \combined{}, at $R \approx 2 \, \um$.

Inspecting the distribution of arrival times as shown in FIG.~\ref{fig:performance-radius}B for $R=2\um$,
we observe a clear difference in skewness between \temporal{} and \spatial{}.
The distribution of arrival times in \spatial{} has long tails since cells that start far away from the source are experiencing very low concentrations of molecules, which disproportionally affect the spatial strategy.
In contrast, \temporal{} shows very few cells that reach the source quickly, as this strategy relies on building memory.
Interestingly, the cells that use \combined{} are both fast and do not get trapped, having both benefits of the other variants.

To evaluate the optimality of the found strategy \combined{}, we compare it against commonly proposed strategies that use memory kernels to integrate temporal information into spatial strategies.
Likewise, we explore a switching strategy in which cells start by using the noise-robust \temporal{} strategy and later switch to using the reactive \spatial{} strategy at a set threshold.
This incorporates the advantages of each variant as shown in Fig~\ref{fig:performance-radius}B.
In all cases, we find that the RL learned strategy outcompetes these simpler explicit strategies (see Appendix \ref{ss:manual}).

\begin{figure}[thb]
    \centering
    \includegraphics[width=\linewidth]{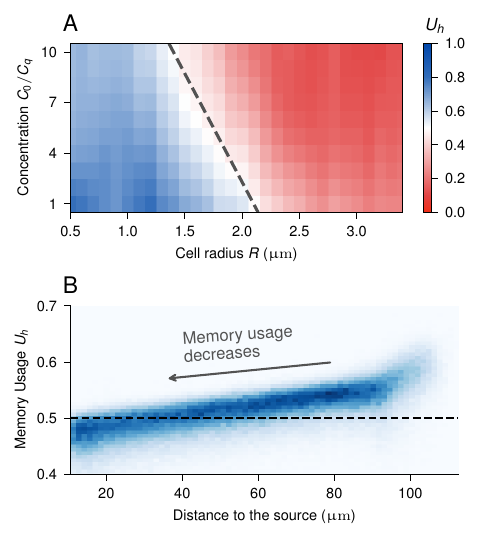}
    \caption{
    \textbf{A} Average memory usage contribution to the steering output during the simulation runs at different sizes and concentration levels $C_0$. 
    The dashed line indicates $U_h \approx 0.5$, i.e. the transition from a memory-dominated strategy to a more reactive sensing-based policy.
    \textbf{B} Distribution of memory usage $U_h$ values during individual trajectories, evaluated at different distances to the source. $R=2 \, \um$. 
    }
    \label{fig:memory-usage}
\end{figure}

\subsection{\label{ss:integration}Integrating temporal and spatial information}
Having established that \combined{} can integrate spatial and temporal information to outcompete both \temporal{} and \spatial{}, we move on to studying the internals of \combined{} directly.
The policy $\pi_\combined$ is a highly non-linear, recursive function which we have parameterized by deep learning neural networks --- this at the cost of lack of interpretability.
Nonetheless, numerous techniques have been developed for gaining insight into the internals of a trained neural network, for instance, by estimating the importance of the input variables.
One of the most elegant techniques to study this attribution problem is the method of integrated gradients (IG) \cite{sundararajanAxiomaticAttributionDeep2017},
which calculates the importance of feature $x_i$ as
\begin{equation}
    \mathcal{I}_i = (x_i - x_i') \int_{0}^{1} \frac{\partial \pi(x' + \alpha (x - x'))}{\partial x_i} \, \mathrm{d}\alpha,
    \label{eq:integrated-gradients}
\end{equation}
where $x'$ is a baseline, which we here simply take to be no input $x' = 0$.
IG is \textit{sensitive} meaning $\mathcal{I}_i$ is non-zero if and only if $x_i$ contribute to the output, and satisfies \textit{completeness} such that the attributions sum to the output, i.e. $a_t = \pi(x) = \pi(0) + \sum_i \mathcal{I}_i$.

We use IG to understand how the cell relies on previous measurements transmitted to it by the hidden state $h_{t-1}$, compared to current measurements $m_t$ from the receptors.
We define $U_h$ as the relative importance of memory,
\begin{equation}
    U_h = \dfrac{\sum_{i \in h} |\mathcal{I}_i|}{\sum_j |\mathcal{I}_j|}.
\end{equation}
A $U_h > 0.5$ value indicates that the hidden state contributes more to the output than the current measurement.
Note that the definitions sums over contributions from all hidden states and all measurements, and is thus virtually independent on e.g. the number of hidden states.

FIG.~\ref{fig:memory-usage}A shows how average memory usage $U_h$ changes as a function of cell size and chemoattractant concentration.
We observe a smooth transition of decreasing contribution of memory as the cell gets larger, in accordance with previous conclusions.
This transition occurs at smaller sizes the higher the concentration.
Interestingly, when evaluating $U_h$ within a single environment (FIG.~\ref{fig:memory-usage}B), we observe a decrease in memory usage as the cell approaches the source.
Thus, the cell is adapting between temporally and spatially dominated strategies during a single trajectory, akin to a continuous version of the discrete switching strategy just considered.

Although the input to the neural network policy $\pi$ is the current measurements $m_t$ and the hidden state $h_{t - 1}$, the output $a_t$ can also be considered a function of all previous measurements $\{m_1, m_2, \cdots, m_t\}$, being processed recursively by a sequence of hidden states,
i.e. $a_t = \pi(m_t, \, h_{t-1}) = \pi(m_t, \, m_{t-1}, \, m_{t-2}, \cdots, \, m_0)$.
Applying Eq.~\eqref{eq:integrated-gradients} in this formulation, we can attribute importance individually to all previous measurements on the current output.

\begin{figure*}[thb]
    \centering
    \includegraphics[width=\linewidth]{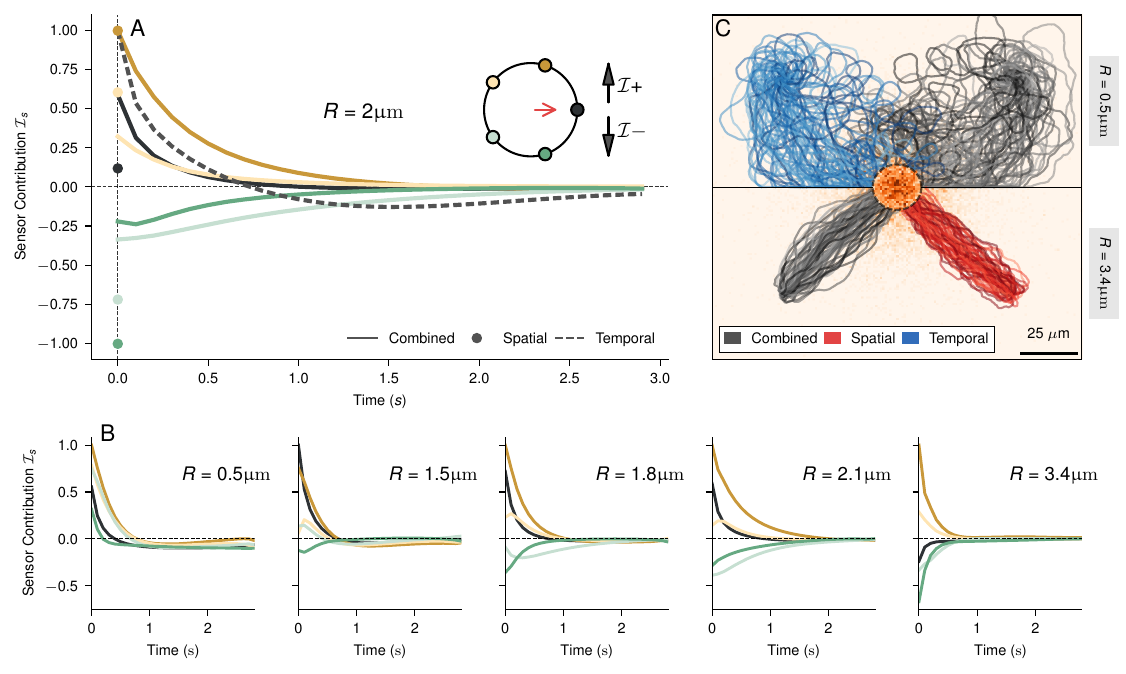}
    \caption{
    \textbf{A}~Contribution of each sensor from past time measurements to the current action. The three variants at $R=2 \, \um $ are shown, with data coloured by sensor position as indicated in the cell diagram.
    For the temporal variant (dashed), only one sensor is shown as all have the same profile as per the designed symmetry.
    The red arrow indicates the swimming direction.
    Curves are obtained by averaging over $\sim 10^5$ trajectories with initial conditions sampled similarly as in previous plots.
    \textbf{B}~Sensor contributions on the combined policy for different cell sizes.
    \textbf{C}~Trajectory visualization of both small (top) and large (bottom) cells.
    See SI for a plot with all three variants.
    }
    \label{fig:integrated-gradients}
\end{figure*}

FIG.~\ref{fig:integrated-gradients}A shows the IG attributions of measurements for the purely spatial, the purely temporal, and the combined strategy at $R = 2 \, \um$.
As the cell diagram indicates, a positive IG value translates into a contribution for a positive reorientation and a negative value vice-versa.
On a pure spatial strategy, the sensors work in opposition, while obviously, there is no contribution from previous measurements.
In contrast, on a pure temporal strategy, all sensors contribute the same, but current measurements are opposed by previous measurements.
Curiously, the shape of the contributions highly resembles the bi-loped shape of the chemotactic memory kernel measured experimentally on the impulse responses of \textit{E. coli} bacteria \cite{celaniBacterialStrategiesChemotaxis2010}.

Similar to the spatial strategy, the combined strategy shows sensors working in opposition, but the left-right symmetry is broken and compensated by temporal variance, with one side dominating early and the other side contributing late.
This sensor signature of the combined strategy makes explicit the non-trivial combination of information it is utilizing, and while these curves are merely IG components, they are indicative of a non-linear combination of information that asymmetrically merges spatial and temporal processing (FIG.~\ref{fig:integrated-gradients}A).
We observe a transition from temporal towards spatial information processing by looking at how measurements are integrated into the combined policy for different sizes (FIG.~\ref{fig:integrated-gradients}B).
This similarity is clearly observed when comparing trajectories of the purely temporal and pure spatial policies with the combined one at the respective extreme cell sizes (FIG.~\ref{fig:integrated-gradients}C).

\section{Discussion}

In this study, we have explored the theoretical possibilities in chemotaxis that arise when traditional limitations are relaxed.
Our findings show that the borders of binary classifications of chemotaxis strategies can be blurred by suitable integration of spatial and temporal information.
In particular, we have shown that for cells with the ability to sense across their bodies as well as having memory access, there is a navigation strategy that outperforms those with only one sensing ability.
Without imposing any constraints on the policy, we have seen the optimal solution to converge to known policies in the limits where it is known that one sensing mechanism clearly provides faster information on the chemical gradient.
Here, we explored this as a function of cell size and found that for large cells, the emerged combined strategy converges to relying only on spatial information, whereas for small microorganisms, the gradient information is strictly obtained on temporal differences.
In the intermediate range, we found no sudden switch in strategy, but instead, the transition between them is continuous and smooth, where information is slowly being integrated by the cell into its decision process.

Our general perspective on chemotaxis is achieved by employing artificial neural networks and optimizing these by reinforcement learning.
The drawback to this is that the obtained strategies are difficult to interpret.
Yet, by comparing analytical strategies and employing integrated gradients to study feature attribution, we find that the optimal strategy that employs both spatial and temporal information is not a simple combination of known strategies, nor is its integration of information types trivial.
Our analysis reveals that memory usage varies with cell size and concentration and changes dynamically throughout trajectories.
This is akin to the well-known phenomenon that cells adapt their measurement sensitivity to local concentration \cite{lazovaResponseRescalingBacterial2011}, but here, we find that in an optimal setting, the navigation strategy itself must also dynamically adapt.

Using DRL to study chemotaxis in the noise-dominated regime is computationally challenging, as it requires a  large number of simulations that must dynamically be run during training.
Our custom approach runs simulations and training on GPU, avoiding slow system-to-device transfers.
Here, we have employed this approach to study a simple chemotactic agent in two dimensions.
An interesting avenue for future research is the move to three dimensions, where the space of possible strategies is qualitatively different.
Likewise, it could be interesting to consider the consequences of a non-static source of chemoattractant or heterogeneous environments and discover the effect of this on a combined chemotactic policy.
Similarly, it is of interest to extend our minimal cell model to specificities of particular organisms, such as a thorough modelling receptor dynamics \cite{aquinoKnowSingleReceptorSensing2016}, the inclusion of stochastic tumbles of peritrichously flagellated bacteria \cite{sourjikRespondingChemicalGradients2012}, or more complex behaviours as the ones seen in \textit{C. elegans}  \cite{iinoParallelUseTwo2009}.

\subsection*{Code availability}

The code for performing training and running the simulations and the scripts to evaluate the strategies can be found at \url{https://github.com/kirkegaardlab/chemoxrl}.

\subsection*{Acknowledgments}
This project has received funding from the Novo Nordisk Foundation Grant Agreement NNF20OC0062047.

\vspace{1cm}
\subsection*{Authors contribution}
A.A. and J.B.K. designed research, performed research, analyzed data, and wrote the paper.

\vspace{1cm}

\appendix*
\section{\label{ss:manual} Comparison with interpretable models}

In this section, we evaluate the chemotactic efficiency of the trained policies by comparing them to simpler strategies where it is straightforward to explain the integration of memory and spatial gradient sensing.

\begin{figure}[h]
    \centering
    \includegraphics[width=\linewidth]{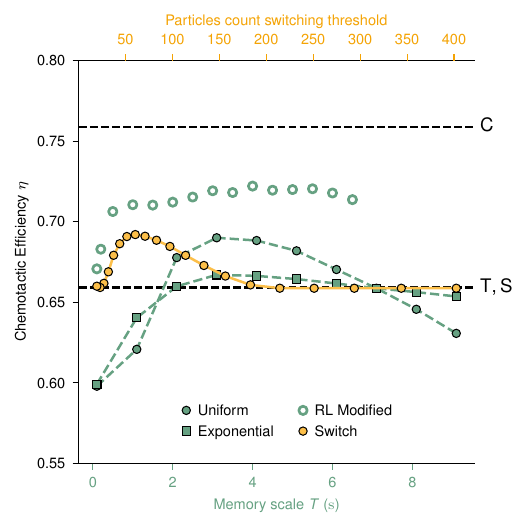}
    \caption{Chemotactic efficiency of proposed explicit policies compared to the neural network policies found using reinforcement learning, at $R=2 \, \um$.
    Green points are for policies that integrate measurements over time (lower axis),
    whereas orange points correspond to the policy achieved by switching between temporal and spatial strategies at a certain concentration threshold (upper axis).
    }
    \label{fig:memory-comparison}
\end{figure}

We begin by defining a naive spatial policy $\tilde{\pi}$ where the steering of the swimming orientation is directly dictated by the strength of the receptor measurements.
The naive optimal reorientation is given by
 \begin{equation}
    \phi = \arctantwo \! {\left( \dfrac{\sin({\boldsymbol \alpha}) \cdot \boldsymbol{\omega}}{\cos({\boldsymbol{\alpha})}\cdot {\boldsymbol{\omega}}} \right)},
 \end{equation}
where ${\boldsymbol \omega}$ are the contributions of each receptor to the decision and ${\boldsymbol \alpha}$ are the angles of the receptor position on the cell's surface with respect to the swimming direction (FIG.~\ref{fig:model-diagram}).
This naive strategy is very susceptible to fluctuations in measurements and can sometimes be improved by restricting the reorientations to a certain $\varepsilon$.
Thus, we consider policies of the form
\begin{equation}
    a_t = \tilde{\pi}(s_t) = \frac{1}{\Delta t} \cdot \begin{cases}
        -\varepsilon & \text{if } \phi \le -\varepsilon \\
        \varepsilon & \text{if } \phi \ge \varepsilon \\
        \phi & \text{otherwise.}
    \end{cases} 
    \label{eq:theta-anal}
\end{equation} 

Integrating measurements over time reduces the fluctuations in concentration measurements, as has also been shown experimentally \cite{endresAccuracyDirectGradient2008}.
Thus, we explore the possibility of cells relying on the average of previous measurements to set the change in orientation.
The contribution of each sensor is then averaged by previous measurements as
\begin{equation} \label{eq:kernel}
    \omega_t^{(i)} = \int_0^{\infty} \kappa(t') \, \hat{m}^{(i)}(t - t') \, \mathrm{d} t'.
\end{equation}
Here, $\hat{m}(t)$ are corrected measurements at time $t$.
Directly using $m(t)$ completely ruins performance, as every time an action is performed, the information of previous measurements is no longer aligned with the cell orientation.
To obtain optimal strategies, we use $\hat{m}(t)$, which is corrected by the action taken $a_t$, and thus only suffers from information decay due to rotational diffusion.

We begin by studying a uniform distribution such as
\begin{equation}
    \kappa(t) = 
        \begin{cases}
        \frac{1}{T} &\text{for } t \le T \\
        0 &\text{otherwise}
        \end{cases}
\end{equation}
where all previous measurements contribute the same up to $T$.
Moreover, we consider the use of an exponentially decaying kernel
\begin{equation}
   \kappa(t) = \frac{1}{T}e^{\nicefrac{-t}{T}} \label{eq:exponential-kernel},
\end{equation}
which gives more weight to newer measurements.

As seen in FIG.~\ref{fig:memory-comparison}, the chemotactic efficiency of these models outperforms \spatial{} and \temporal{} when some rudimentary use of memory is allowed. 
We note that each reported value on the analytical strategies is evaluated with different $\varepsilon$ and only the best-performing one is shown. 
Nevertheless, we observe that a large memory timescale becomes counterproductive as the movement of the cell makes previous measurements irrelevant and only contributes noise to the decision.
Despite the gain in efficiency, the optimal timescale for the proposed models is far from reaching the chemotactic efficiency of \combined{}.

We note that as $T \to 0$, \spatial{} outperforms the explicit models.
This can be explained by the freedom of \spatial{} to dynamically control a non-linear equivalent of $\varepsilon$ depending on the measurements.
With this in mind, we investigate a new RL agent using the same neural network as \spatial{}, but whose input is given by Eq. \eqref{eq:kernel}.
Thus the integration of memory is fully controlled, but any non-linear action can be taken based on this input.
We note that this again requires correcting previous inputs and special attention is given to the early parts of trajectories, such that the policy only averages over known measurements.
FIG.~\ref{fig:memory-comparison} shows that this indeed outperforms \spatial{} and \temporal{}, but cannot reach the performance of \combined{}. 
This suggests that \combined{} is not just combining a \textit{temporal average} with a \textit{spatial strategy} but is also using elements of a \textit{temporal strategy}.

Spermatozoa have recently been shown to exhibit a biphasic chemotactic strategy, in which there is a concentration-dependent switch between hyperactive phases, characterized by random changes in orientation, and more well-known chiral motion \cite{zaferaniBiphasicChemokinesisMammalian2023}.
Presently, a switch between a temporal and a spatial strategy could achieve the best of the distinct time distributions of \temporal{} and \spatial{} in FIG.~\ref{fig:performance-radius}B.
We implement this by setting a cutoff particle count at which we switch from \temporal{} to \spatial{}.
As a function of this threshold, an increase in chemotactic efficiency is observed, as shown in FIG.~\ref{fig:memory-comparison}, but this also does not reach the efficiency achieved by \combined{}.
Nevertheless, the increase in efficiency does suggest that the contribution of temporal and spatial may change dynamically with the concentrations.

Finally, we explore the possibility of designing an agent where the effective memory scale $T$ is linearly dependent on the measurement concentration, as suggested in FIG.~\ref{fig:memory-usage}B, such that
\begin{equation}
    T = A \ \langle m_t \rangle + B
\end{equation}

\begin{figure}[h]
    \centering
    \includegraphics[width=3.3in]{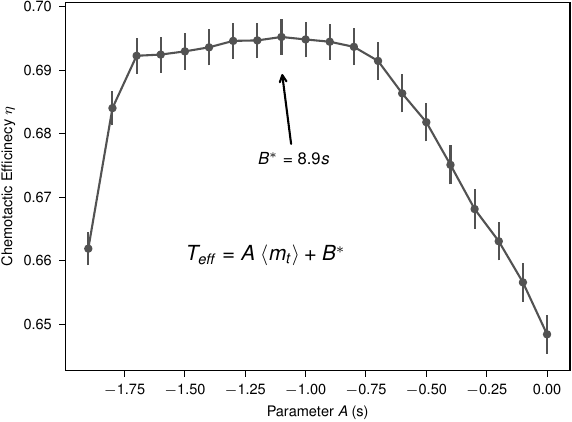}
    \caption{Chemotactic efficiency of a policy that adjusts the memory time scale according to a linear dependency with the average strength of the measurements $T=A\langle m_t \rangle + B$. The efficiency shown as a function of $A$, and $B$ and $\varepsilon$ are the optimal values for that $A$. The simulations parameters are the same as in FIG.~\ref{fig:memory-comparison}.}
    \label{fig:linear-memory}
\end{figure}

We evaluate for different parameters of $A$, $B$, and $\varepsilon$ on a uniform kernel.
FIG~\ref{fig:linear-memory} shows the chemotactic efficiency at different parameters $A$, with the best performant $B^\ast(A)$ and $\varepsilon^\ast(A)$.

The performance of this model is similar to that of a fixed uniform kernel.
While the study of integrated gradients shows the amount of memory used, it does not reveal how this memory is used, and in particular here we find that a simple uniform kernel is far from enough to reach optimal behavior.    

\bibliography{references}

\end{document}